\documentclass[conference]{IEEEtran}

\usepackage{cite}
\usepackage[T1]{fontenc}
\usepackage{amsmath,amssymb,amsfonts}
\usepackage{graphicx}
\usepackage{placeins}
\usepackage{textcomp}
\usepackage{xcolor}

\begin{document}

\title{Modular Inverse Dynamics and Wrench Recovery for Closed-Chain Mechanisms Using Path Assembly}

\author{

\IEEEauthorblockN{Mohammad Dastranj\textsuperscript{*}}
\IEEEauthorblockA{
\textit{Unit of Automation Technology}\\
\textit{and Mechanical Engineering}\\
\textit{Tampere University}\\
Tampere, Finland\\
mohammad.dastranj@tuni.fi\\
\textsuperscript{*}Corresponding author
}

\and

\IEEEauthorblockN{Jouni Mattila}
\IEEEauthorblockA{
\textit{Unit of Automation Technology}\\
\textit{and Mechanical Engineering}\\
\textit{Tampere University}\\
Tampere, Finland\\
jouni.mattila@tuni.fi
}
}

\maketitle


\begin{abstract}
Rigid closed chains can admit indeterminate internal reactions
even when their motion is fully specified. This paper develops
a modular inverse-dynamics workflow that combines path-based
motion reconstruction with recovery of uniquely determined
actuator and joint-wrench outputs. Analytical second-order
differentiation of assembled transformation paths supplies
passive velocities and accelerations on a regular closure branch.
Spatial Newton--Euler balances are then assembled over the
physical body-and-joint graph. Established nullspace and dual
criteria identify and recover unique outputs, while balance,
ideal-joint admissibility and affine-family diagnostics assess
nonunique reference reactions. Evaluation against Simscape
Multibody uses prescribed motion of a seven-degree-of-freedom
heavy-duty hybrid serial--parallel manipulator under common
rigid-body, ideal-joint assumptions. The results show close agreement with the reference for passive positions, actuator efforts and uniquely recoverable constraint-wrench components. Reaction diagnostics support compatibility with the admissible numerical load family.
\end{abstract}

\begin{IEEEkeywords}
Closed-chain mechanisms, modular modeling, inverse dynamics,
path assembly, joint-wrench recovery.
\end{IEEEkeywords}


\section{Introduction}
\label{sec:introduction}

Modular modeling supports the reuse of submechanism descriptions
while preserving their kinematic and dynamic coupling.
Systematic treatments of closed chains include modular dynamics
for parallel manipulators with hybrid limbs~\cite{muller2022}
and constraint embedding for recursive dynamics with local
loops~\cite{chignoli2025}.
Virtual Decomposition Control (VDC) likewise requires compatible
motion and wrench quantities for its virtual-power-flow
analysis~\cite{hejrati2024};
applications include heavy-duty hydraulic
manipulators~\cite{petrovic2022,hejrati2025}
and, more recently, hydraulic
robot legs with triangular closed chains~\cite{zhang2025}.
These developments motivate reusable motion and load descriptions
for closed-chain modules.

Closure-consistent motion, however, does not imply that every
joint load is uniquely determined. Redundant rigid constraints
can admit multiple reaction distributions while selected
components remain unique~\cite{wojtyra2005}.
Free-body-diagram~\cite{pekal2019}
and constraint-matrix~\cite{pekal2023} approaches have been
developed for analyzing reaction and driving-force uniqueness,
with subsequent
developments within the divide-and-conquer
framework~\cite{pekal2026}.
Consequently, load recovery requires identifying which requested
outputs are invariant across the admissible solutions of the
balance equations.

Path-Assembled Closure Differential Mapping (PACDM) provides
closure resolution and local active-to-passive mappings from
ordered transformation paths~\cite{dastranj2026}.
This representation has also been connected to actuator-port
mappings, acceleration reduction, and reduced inverse and forward
dynamics~\cite{shahna2026}. However, integrating reaction-uniqueness analysis into PACDM-based inverse dynamics for physical actuator and joint-wrench recovery remains to be demonstrated.
Building on these developments and established body-balance and
reaction-uniqueness formulations~\cite{pekal2019,pekal2023,pekal2025},
the present work develops a path-assembly workflow for motion
reconstruction and recovery of uniquely determined actuator and
joint-wrench outputs in the presence of indeterminate internal
reactions. Here, modularity refers to module-local closure reconstruction and assembly of body-level balance equations; the coupled load-recovery problem is solved globally.

The workflow extends the inherited kinematics through analytical
second-order path differentiation, reconstructing passive
velocities and accelerations on a regular closure branch.
Spatial Newton--Euler balances are assembled over an explicitly
specified physical body-and-joint graph, with each body's
inertia included once. The established nullspace criterion
and its dual formulation recover each determined output,
subject to numerical consistency and rank checks.
Reference reactions that are nonunique are assessed through
body balance, ideal-joint admissibility and distance to the
normalized affine solution family.
Together, these elements retain the modular path description
while separating uniquely recoverable loads from indeterminate
reaction distributions.

The formulation is evaluated against Simscape Multibody under
prescribed motion of a 7-degree-of-freedom heavy-duty
hybrid serial--parallel manipulator with two closed-chain modules.
Simulation provides reference constraint forces and torques that
are difficult to measure comprehensively in
hardware~\cite{pekal2025}.
The common rigid-body, ideal-joint setting isolates reconstruction
and load recovery from measurement noise and unmodeled physical
disturbances.


\section{Path-Assembly Kinematics}
\label{sec:kinematics}

\subsection{Problem Formulation and Conventions}
\label{subsec:problem-conventions}

The reconstruction problem concerns the prescribed motion of a
rigid closed-chain manipulator with ideal revolute and prismatic
joints. The active trajectory and its first two time derivatives
are given, while the passive motion is determined by the closure
equations. The active and passive subvectors of the full
coordinate vector $\boldsymbol{q}\in\mathbb{R}^{n}$ are defined by
\begin{equation}
    \boldsymbol{q}_{a}=\boldsymbol{P}_{a}\boldsymbol{q},
    \qquad
    \boldsymbol{q}_{p}=\boldsymbol{P}_{p}\boldsymbol{q},
    \label{eq:coordinate-selection}
\end{equation}
where the constant selection matrices $\boldsymbol{P}_{a}$ and
$\boldsymbol{P}_{p}$ extract the $n_a$ active and $n_p$ passive
coordinates, respectively, with $n_a+n_p=n$. The prescribed
trajectory $\boldsymbol{q}_{a}(t)$ is assumed to be twice
continuously differentiable. Reconstruction is restricted to a
selected regular closure branch on which the passive coordinates
are locally determined by the active coordinates.

Following the path-assembly representation
of~\cite{dastranj2026}, each closed-chain module is described by
ordered transformation paths sharing the same endpoints. These
paths comprise fixed frame transformations and constant-twist
joint transformations. The module partition, path definitions
and geometry are specified inputs to the reconstruction.

The transformation $\boldsymbol{T}_{AB}\in SE(3)$ maps
coordinates from frame $B$ to frame $A$. The wedge and vee maps between
$\mathbb{R}^{6}$ and $\mathfrak{se}(3)$ use angular-first
vector ordering, with a twist expressed as
$\boldsymbol{V}=\operatorname{col}(\boldsymbol{\omega},
\boldsymbol{v})$, where $\operatorname{col}(\cdot)$ denotes
vertical concatenation. Revolute and prismatic coordinates are
expressed in radians and metres, respectively. The reconstructed
positions, velocities and accelerations provide the kinematic
inputs to the physical body balances in Section~\ref{sec:dynamics}.
These balances are subsequently used to assess the uniqueness
of the requested actuator efforts and joint-wrench components
and to recover the uniquely determined outputs.

\subsection{Analytical Motion Reconstruction}
\label{subsec:motion-reconstruction}

The closure representation and local differential mapping
of~\cite{dastranj2026} provide the basis for motion reconstruction.
For each independently resolved closed-chain module $m$, the
$n_{m,a}$ active and $n_{m,p}$ passive coordinates form the vector
$\widehat{\boldsymbol{q}}_m=\nobreak\operatorname{col}
(\widehat{\boldsymbol{q}}_{m,a},\widehat{\boldsymbol{q}}_{m,p})$.
For two paths $\boldsymbol{T}_{\ell}^{\langle+\rangle}$ and
$\boldsymbol{T}_{\ell}^{\langle-\rangle}$ sharing endpoints, closure is
expressed within the selected local logarithm chart as
\begin{equation}
    \boldsymbol{\rho}_{\ell}
    =\left[\operatorname{Log}\!\left(
    (\boldsymbol{T}_{\ell}^{\langle-\rangle})^{-1}
    \boldsymbol{T}_{\ell}^{\langle+\rangle}\right)\right]^{\vee}
    =\boldsymbol{0}_{6}.
    \label{eq:path-closure}
\end{equation}
For a module with $N_m$ paths, path~1 is retained as the
reference $\boldsymbol{T}_{\ell}^{\langle+\rangle}$ in every
comparison, while $\boldsymbol{T}_{\ell}^{\langle-\rangle}$
runs through the remaining paths. The resulting $N_m-1$
residuals are assembled into
$\overline{\boldsymbol{\rho}}_m\in\mathbb{R}^{6(N_m-1)}$.
Although these residuals enforce equality of all path
transformations, their scalar components may be dependent.
The corresponding module-coordinate Jacobian is
$\overline{\boldsymbol{R}}_m
=\partial\overline{\boldsymbol{\rho}}_m/
\partial\widehat{\boldsymbol{q}}_m$.
A row-selection matrix $\boldsymbol{S}_m$ retains $n_{m,p}$
equations, giving
\begin{equation}
    \begin{aligned}
        \boldsymbol{c}_m
        &=\boldsymbol{S}_m\overline{\boldsymbol{\rho}}_m
        =\boldsymbol{0},\\
        \boldsymbol{R}_m
        &=\boldsymbol{S}_m\overline{\boldsymbol{R}}_m
        =\begin{bmatrix}
            \boldsymbol{R}_{m,a} & \boldsymbol{R}_{m,p}
        \end{bmatrix}.
    \end{aligned}
    \label{eq:selected-closure}
\end{equation}
Local reconstruction assumes that
$\overline{\boldsymbol{R}}_m$ has constant rank $n_{m,p}$
in a neighborhood of the considered configuration and that
the selected passive block $\boldsymbol{R}_{m,p}$ is nonsingular.
Under these conditions, the selected equations are locally
equivalent to the full closure equations and determine the
passive coordinates from the active coordinates.
Differentiation of~\eqref{eq:selected-closure} gives the passive
coordinate rates. Assembly of the module maps in the global
coordinate order then yields
\begin{equation}
    \begin{aligned}
        \dot{\widehat{\boldsymbol{q}}}_{m,p}
        &=-\boldsymbol{R}_{m,p}^{-1}\boldsymbol{R}_{m,a}
        \dot{\widehat{\boldsymbol{q}}}_{m,a},\\
        \dot{\boldsymbol{q}}
        &=\boldsymbol{E}(\boldsymbol{q})\dot{\boldsymbol{q}}_a,
        \qquad
        \boldsymbol{P}_a\boldsymbol{E}
        =\boldsymbol{I}_{n_a},
    \end{aligned}
    \label{eq:velocity-reconstruction}
\end{equation}
where the intermediate Jacobian
$\boldsymbol{E}\in\mathbb{R}^{n\times n_a}$ maps the
prescribed active rates to the full coordinate rates.

A closure-consistent initial configuration is obtained by
defect homotopy as in~\cite{dastranj2026}. The present
implementation retains the row-selection matrix
$\boldsymbol{S}_m$ established at initialization.
At subsequent trajectory points, the previously accepted
configuration provides a first-order prediction, followed by
Newton correction with passive-coordinate bounds and line
search. The corrected configuration is checked against the
full closure residual.

The corresponding acceleration follows by time differentiation
of~\eqref{eq:velocity-reconstruction}, in accordance with the
standard dependent-coordinate acceleration
relation~\cite{muller2019},
\begin{equation}
    \begin{aligned}
        \ddot{\boldsymbol{q}}
        &=\boldsymbol{E}\ddot{\boldsymbol{q}}_a
        +\boldsymbol{\gamma},\\
        \boldsymbol{\gamma}
        &=\dot{\boldsymbol{E}}\dot{\boldsymbol{q}}_a,
        \qquad
        \boldsymbol{P}_a\boldsymbol{\gamma}=\boldsymbol{0}.
    \end{aligned}
    \label{eq:motion-map}
\end{equation}
The acceleration bias $\boldsymbol{\gamma}$ accounts for the
configuration dependence of the intermediate Jacobian. Its
evaluation uses analytical derivatives of the path factors.
A joint factor $\boldsymbol{F}(q_j)$ has a constant generator
$\boldsymbol{X}\in\mathfrak{se}(3)$, including any signed
coordinate scale, which satisfies
$\partial\boldsymbol{F}/\partial q_j=\boldsymbol{X}\boldsymbol{F}$~\cite{muller2019}.
Its first and second time derivatives are therefore
\begin{equation}
    \begin{aligned}
        \dot{\boldsymbol{F}}
        &=\boldsymbol{X}\boldsymbol{F}\dot{q}_j,\\
        \ddot{\boldsymbol{F}}
        &=\boldsymbol{X}^{2}\boldsymbol{F}\dot{q}_j^{2}
        +\boldsymbol{X}\boldsymbol{F}\ddot{q}_j.
    \end{aligned}
    \label{eq:factor-derivatives}
\end{equation}
For a fixed factor, both derivatives vanish. When a factor
$\boldsymbol{F}$ is appended to a partial path product
$\boldsymbol{T}$, the product rule gives
\begin{equation}
    \begin{aligned}
        \frac{d}{dt}(\boldsymbol{T}\boldsymbol{F})
        &=\dot{\boldsymbol{T}}\boldsymbol{F}
        +\boldsymbol{T}\dot{\boldsymbol{F}},\\
        \frac{d^2}{dt^2}(\boldsymbol{T}\boldsymbol{F})
        &=\ddot{\boldsymbol{T}}\boldsymbol{F}
        +2\dot{\boldsymbol{T}}\dot{\boldsymbol{F}}
        +\boldsymbol{T}\ddot{\boldsymbol{F}}.
    \end{aligned}
    \label{eq:path-derivatives}
\end{equation}
Successive application of these relations provides the first
and second derivatives of each assembled path.

The closure acceleration bias is isolated by evaluating these
recursions at the reconstructed $\boldsymbol{q}$ and
$\dot{\boldsymbol{q}}$ with a zero acceleration argument.
The resulting second path derivatives are denoted by
$\ddot{\boldsymbol{T}}_{\ell,0}^{\langle\pm\rangle}$.
At closure and for a tangent velocity, the relative path transform
and its first derivative are $\boldsymbol{I}_4$ and zero,
respectively. The logarithm differential used in
PACDM~\cite{dastranj2026} reduces to the identity
at this point. Since the first derivative of the relative
transform vanishes, the second derivative of the logarithmic
residual gives the closure acceleration bias as
\begin{equation}
    \boldsymbol{\beta}_{\ell}
    =\left[
        (\boldsymbol{T}_{\ell}^{\langle-\rangle})^{-1}
        \left(
            \ddot{\boldsymbol{T}}_{\ell,0}^{\langle+\rangle}
            -\ddot{\boldsymbol{T}}_{\ell,0}^{\langle-\rangle}
        \right)
    \right]^{\vee}.
    \label{eq:closure-bias}
\end{equation}
The module bias vector $\overline{\boldsymbol{\beta}}_m$
contains these pairwise terms in the same order as the residuals
in $\overline{\boldsymbol{\rho}}_m$. The corresponding
acceleration constraints follow by twice differentiating the
selected closure equations~\cite{muller2019}.
Together with~\eqref{eq:motion-map}, they give
\begin{equation}
    \begin{aligned}
        \boldsymbol{R}_m\ddot{\widehat{\boldsymbol{q}}}_m
        +\boldsymbol{S}_m\overline{\boldsymbol{\beta}}_m
        &=\boldsymbol{0},\\
        \boldsymbol{\gamma}_{m,p}
        &=-\boldsymbol{R}_{m,p}^{-1}
        \boldsymbol{S}_m\overline{\boldsymbol{\beta}}_m,
    \end{aligned}
    \label{eq:acceleration-bias-solve}
\end{equation}
where $\boldsymbol{\gamma}_{m,p}$ collects the passive entries
of $\boldsymbol{\gamma}$ for module $m$. Since the active
entries are zero, assembly of these module solutions determines
$\boldsymbol{\gamma}$ and hence the full coordinate acceleration
in~\eqref{eq:motion-map}. The reconstructed motion is checked
through the full velocity closure equations and the full
acceleration mismatch matrices of the path pairs.


\section{Inverse Dynamics and Wrench Recovery}
\label{sec:dynamics}

\subsection{Physical Inverse-Dynamics Assembly}
\label{subsec:physical-assembly}

The reconstructed coordinate motion determines the pose and
motion of every physical body. For body $i$,
$\boldsymbol{T}_i=\boldsymbol{T}_{0i}$ maps its body reference
frame to the fixed world frame, and its derivatives are evaluated
from the assigned path using~\eqref{eq:path-derivatives}.
The body twist and its time derivative follow from the standard
body-twist definition~\cite{lynch2017} as
\begin{equation}
    \begin{aligned}
        \boldsymbol{V}_i
        &=\left(\boldsymbol{T}_i^{-1}
        \dot{\boldsymbol{T}}_i\right)^{\vee},\\
        \dot{\boldsymbol{V}}_i
        &=\left[
            \boldsymbol{T}_i^{-1}\ddot{\boldsymbol{T}}_i
            -\left(\boldsymbol{T}_i^{-1}
            \dot{\boldsymbol{T}}_i\right)^2
        \right]^{\vee}.
    \end{aligned}
    \label{eq:body-motion}
\end{equation}
Both quantities are expressed in body axes. The derivative
$\dot{\boldsymbol{V}}_i$ is the ordinary time derivative of
these body-frame components, so its linear part includes the
effect of the rotating axes.

The constant spatial inertia
$\boldsymbol{\mathcal{I}}_i\in\mathbb{R}^{6\times6}$ is
expressed about the body reference origin in body axes and
represents its mass distribution~\cite{lynch2017}.
Wrenches use moment-first ordering, with moments in
$\mathrm{N.m}$ and forces in $\mathrm{N}$.
For $\boldsymbol{V}_i=\operatorname{col}
(\boldsymbol{\omega}_i,\boldsymbol{v}_i)$, the dual spatial
cross-product operator is
\begin{equation}
    \boldsymbol{V}_i\times^{*}
    =\begin{bmatrix}
        [\boldsymbol{\omega}_i]_{\times}
        & [\boldsymbol{v}_i]_{\times}\\
        \boldsymbol{0}_{3\times3}
        & [\boldsymbol{\omega}_i]_{\times}
    \end{bmatrix},
    \label{eq:force-cross-product}
\end{equation}
where $[\boldsymbol{x}]_{\times}\boldsymbol{y}
=\boldsymbol{x}\times\boldsymbol{y}$.
The spatial Newton--Euler balance~\cite{featherstone2000}
then gives the net wrench $\boldsymbol{b}_i$ required from
the incident joints as
\begin{equation}
    \boldsymbol{b}_i
    =\boldsymbol{\mathcal{I}}_i\dot{\boldsymbol{V}}_i
    +(\boldsymbol{V}_i\times^{*})
    \boldsymbol{\mathcal{I}}_i\boldsymbol{V}_i
    -\boldsymbol{w}_{g,i}-\boldsymbol{w}_{\mathrm{ext},i}.
    \label{eq:body-balance}
\end{equation}
Gravity contributes
$\boldsymbol{w}_{g,i}=\operatorname{col}
(\boldsymbol{c}_i\times\boldsymbol{f}_{g,i},
\boldsymbol{f}_{g,i})$, with
$\boldsymbol{f}_{g,i}=\nobreak m_i\boldsymbol{R}_i^{T}\boldsymbol{g}$.
Here, $m_i$ is the body mass, $\boldsymbol{c}_i$ is its
center-of-mass offset from the body origin in body axes,
$\boldsymbol{R}_i$ is the rotation block of $\boldsymbol{T}_i$,
and $\boldsymbol{g}$ is the gravitational acceleration in
world axes. The term $\boldsymbol{w}_{\mathrm{ext},i}$ denotes
a known nongravitational wrench applied to body $i$, expressed
about its reference origin in body axes. Nonzero prescribed
external loads are included directly through this term.
One balance is assembled for each physical moving body or
rigid aggregate using its total spatial inertia.

Joint $j$ connects a base body $a_j$ to a follower body $f_j$,
with $a_j=0$ denoting the world.
Its total wrench $\boldsymbol{w}_j$ acts from base to follower
and is expressed about the origin of the joint follower frame
$F_j$ in its axes. For an ideal revolute or prismatic joint,
the reaction basis $\boldsymbol{C}_j\in\mathbb{R}^{6\times5}$
has orthonormal columns satisfying
$\boldsymbol{S}_j^T\boldsymbol{C}_j=\boldsymbol{0}^T$,
where $\boldsymbol{S}_j\in\mathbb{R}^{6}$ is the unit-axis
motion screw expressed in $F_j$.
This reciprocity expresses the workless character of ideal
constraints~\cite{lynch2017}.
The joint wrench therefore has the form
\begin{equation}
    \boldsymbol{w}_j
    =\boldsymbol{C}_j\boldsymbol{\lambda}_j
    +\boldsymbol{U}_j u_j,
    \label{eq:joint-wrench}
\end{equation}
where $\boldsymbol{\lambda}_j\in\mathbb{R}^{5}$ contains
the reaction coordinates. For an active joint,
$\boldsymbol{U}_j=\boldsymbol{S}_j$ and $u_j$ is the
actuator torque or force; passive joints have no actuation
term or associated load column.
The effort sign follows the physical positive joint axis,
independently of path traversal direction.

For a transform $\boldsymbol{T}$ with rotation
$\boldsymbol{R}$ and translation $\boldsymbol{p}$,
the corresponding wrench transport is~\cite{lynch2017}
\begin{equation}
    \boldsymbol{\mathcal{W}}(\boldsymbol{T})
    =\begin{bmatrix}
        \boldsymbol{R} & [\boldsymbol{p}]_{\times}\boldsymbol{R}\\
        \boldsymbol{0}_{3\times3} & \boldsymbol{R}
    \end{bmatrix}.
    \label{eq:wrench-transport}
\end{equation}
The off-diagonal block accounts for the change of moment
reference. The load vector
$\boldsymbol{\chi}\in\mathbb{R}^{n_\chi}$ contains the
reaction coordinates and active efforts in joint order.
The blocks of $\boldsymbol{A}$ associated with joint $j$
and its two incident bodies are
\begin{equation}
    \begin{aligned}
        \boldsymbol{A}_{f_j,j}
        &=\boldsymbol{\mathcal{W}}(\boldsymbol{T}_{f_jF_j})
        [\boldsymbol{C}_j\ \boldsymbol{U}_j],\\
        \boldsymbol{A}_{a_j,j}
        &=-\boldsymbol{\mathcal{W}}(\boldsymbol{T}_{a_jF_j})
        [\boldsymbol{C}_j\ \boldsymbol{U}_j].
    \end{aligned}
    \label{eq:joint-assembly}
\end{equation}
Here, $\boldsymbol{T}_{iF_j}$ maps the joint follower frame
to body frame $i$. Both terms transport the same wrench
reference, with opposite signs for action and reaction.
The base block is omitted for $a_j=0$, and all other body
blocks in these joint columns are zero.
Assembly over the moving bodies gives
\begin{equation}
    \boldsymbol{A}\boldsymbol{\chi}=\boldsymbol{b},
    \qquad
    \boldsymbol{y}=\boldsymbol{L}\boldsymbol{\chi},
    \label{eq:assembled-balance}
\end{equation}
where $\boldsymbol{b}=\operatorname{col}
(\boldsymbol{b}_1,\ldots,\boldsymbol{b}_{n_b})$ contains
the wrench requirements in~\eqref{eq:body-balance}.
For $n_b$ moving bodies and $n_y$ scalar outputs,
$\boldsymbol{A}\in\mathbb{R}^{6n_b\times n_\chi}$ and
$\boldsymbol{L}\in\mathbb{R}^{n_y\times n_\chi}$.
Rows of $\boldsymbol{L}$ select actuator efforts or components
of the constraint wrench $\boldsymbol{C}_j\boldsymbol{\lambda}_j$
or total wrench $\boldsymbol{w}_j$, after transport to the
specified output frame. Thus, each output retains its joint,
frame, component and physical meaning.

\subsection{Output Uniqueness and Recovery}
\label{subsec:output-recovery}

Output uniqueness follows the established nullspace criterion
for reaction components and their linear
combinations~\cite{pekal2025}.
For a consistent balance system, a particular solution
$\boldsymbol{\chi}_0$ and a basis
$\boldsymbol{N}_\chi$ of $\ker(\boldsymbol{A})$
describe all solutions as
\begin{equation}
    \boldsymbol{\chi}
    =\boldsymbol{\chi}_0+\boldsymbol{N}_\chi\boldsymbol{\eta},
    \label{eq:load-solution-family}
\end{equation}
where $\boldsymbol{\eta}$ is arbitrary.
For row $\boldsymbol{\ell}_k^T$ of $\boldsymbol{L}$,
the corresponding output $y_k$ is independent of this freedom
if and only if
\begin{equation}
    \begin{aligned}
        \boldsymbol{\ell}_k^T\boldsymbol{N}_\chi
        &=\boldsymbol{0}^T\\
        &\Longleftrightarrow\
        \boldsymbol{\ell}_k
        \in\operatorname{range}(\boldsymbol{A}^T)\\
        &\Longleftrightarrow\
        \exists\,\boldsymbol{d}_k:
        \boldsymbol{A}^T\boldsymbol{d}_k
        =\boldsymbol{\ell}_k .
    \end{aligned}
    \label{eq:output-uniqueness}
\end{equation}
The first condition removes the term
$\boldsymbol{\ell}_k^T\boldsymbol{N}_\chi\boldsymbol{\eta}$,
and the remaining equivalences follow from
$\ker(\boldsymbol{A})^\perp
=\operatorname{range}(\boldsymbol{A}^T)$.
Consequently, any such dual vector
$\boldsymbol{d}_k\in\mathbb{R}^{6n_b}$ recovers the output as
\begin{equation}
    y_k=\boldsymbol{d}_k^T\boldsymbol{b}.
    \label{eq:dual-recovery}
\end{equation}
Thus, selected outputs can be determined even when the
complete load vector is nonunique.

The numerical calculation uses positive diagonal scales
$\boldsymbol{S}_e$, $\boldsymbol{S}_x$ and
$\boldsymbol{S}_y$ for the balance equations, load coordinates
and outputs, respectively:
\begin{equation}
    \begin{aligned}
        \boldsymbol{A}_n
        &=\boldsymbol{S}_e^{-1}\boldsymbol{A}\boldsymbol{S}_x,
        &\boldsymbol{b}_n&=\boldsymbol{S}_e^{-1}\boldsymbol{b},\\
        \boldsymbol{L}_n
        &=\boldsymbol{S}_y^{-1}\boldsymbol{L}\boldsymbol{S}_x,
        &\boldsymbol{z}&=\boldsymbol{S}_x^{-1}\boldsymbol{\chi}.
    \end{aligned}
    \label{eq:normalized-system}
\end{equation}
These scales make the numerical tests dimensionless.
A full singular value decomposition
$\boldsymbol{A}_n=\boldsymbol{U}\boldsymbol{\Sigma}
\boldsymbol{V}^T$~\cite{pekal2025} provides the retained factors
$\boldsymbol{U}_r$, $\boldsymbol{\Sigma}_r$ and
$\boldsymbol{V}_r$ for numerical rank $r$.
The remaining right singular vectors form
$\boldsymbol{N}=[\boldsymbol{v}_{r+1},\ldots,
\boldsymbol{v}_{n_\chi}]$.
The complete right basis is retained because the physical
balance matrix may have more columns than rows.
The representative solution and dual recovery are
\begin{equation}
    \begin{aligned}
        \boldsymbol{z}_0
        &=\boldsymbol{V}_r\boldsymbol{\Sigma}_r^{-1}
        \boldsymbol{U}_r^T\boldsymbol{b}_n,\\
        \boldsymbol{D}_n
        &=\boldsymbol{U}_r\boldsymbol{\Sigma}_r^{-1}
        \boldsymbol{V}_r^T\boldsymbol{L}_n^T,\\
        \boldsymbol{y}
        &=\boldsymbol{S}_y\boldsymbol{D}_n^T\boldsymbol{b}_n,
    \end{aligned}
    \label{eq:svd-recovery}
\end{equation}
with the last line providing candidate output values.

Acceptance is assessed separately for each output.
For row $\boldsymbol{\ell}_{n,k}^T$ of
$\boldsymbol{L}_n$ and column $\boldsymbol{d}_{n,k}$
of $\boldsymbol{D}_n$, both
$\|\boldsymbol{\ell}_{n,k}^T\boldsymbol{N}\|_2$ and
$\|\boldsymbol{A}_n^T\boldsymbol{d}_{n,k}
-\boldsymbol{\ell}_{n,k}\|_2$ must satisfy the prescribed
absolute and row-relative tolerances.
Acceptance also requires a sufficiently small normalized
balance residual
$\|\boldsymbol{A}_n\boldsymbol{z}_0-\boldsymbol{b}_n\|_\infty$
and an unchanged numerical rank when the singular-value
threshold is multiplied and divided by ten.
Outputs failing any of these conditions are marked unavailable.
The representative $\boldsymbol{z}_0$ is retained for
diagnostics; its undetermined load components are not
reported as unique predictions.

\begin{figure*}[tb]
    \centering
    \includegraphics[width=0.44\textwidth,
        trim=0 0 0 0,clip]{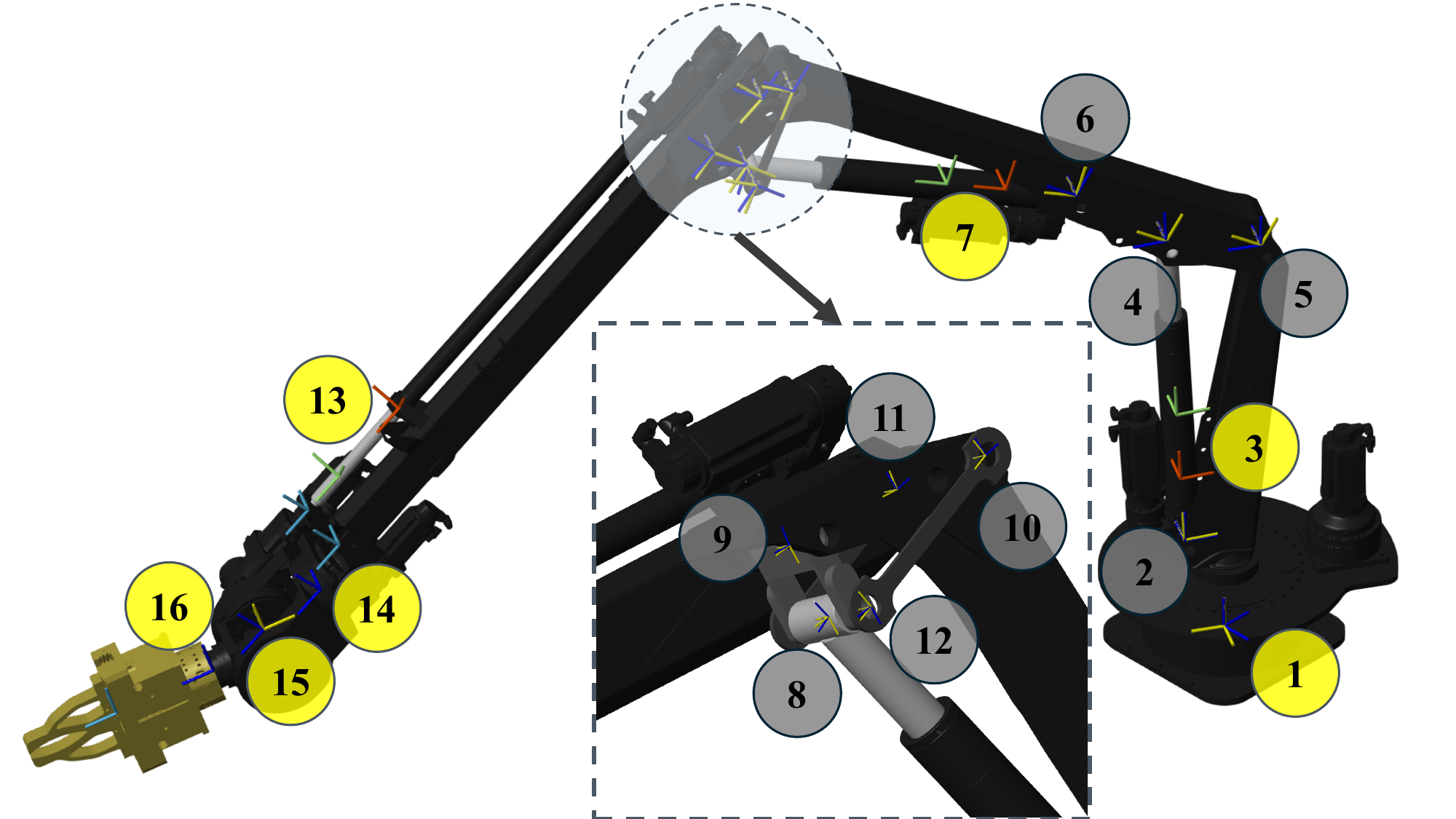}\hfill
    \includegraphics[width=0.44\textwidth,
        trim=0 0 0 0,clip]{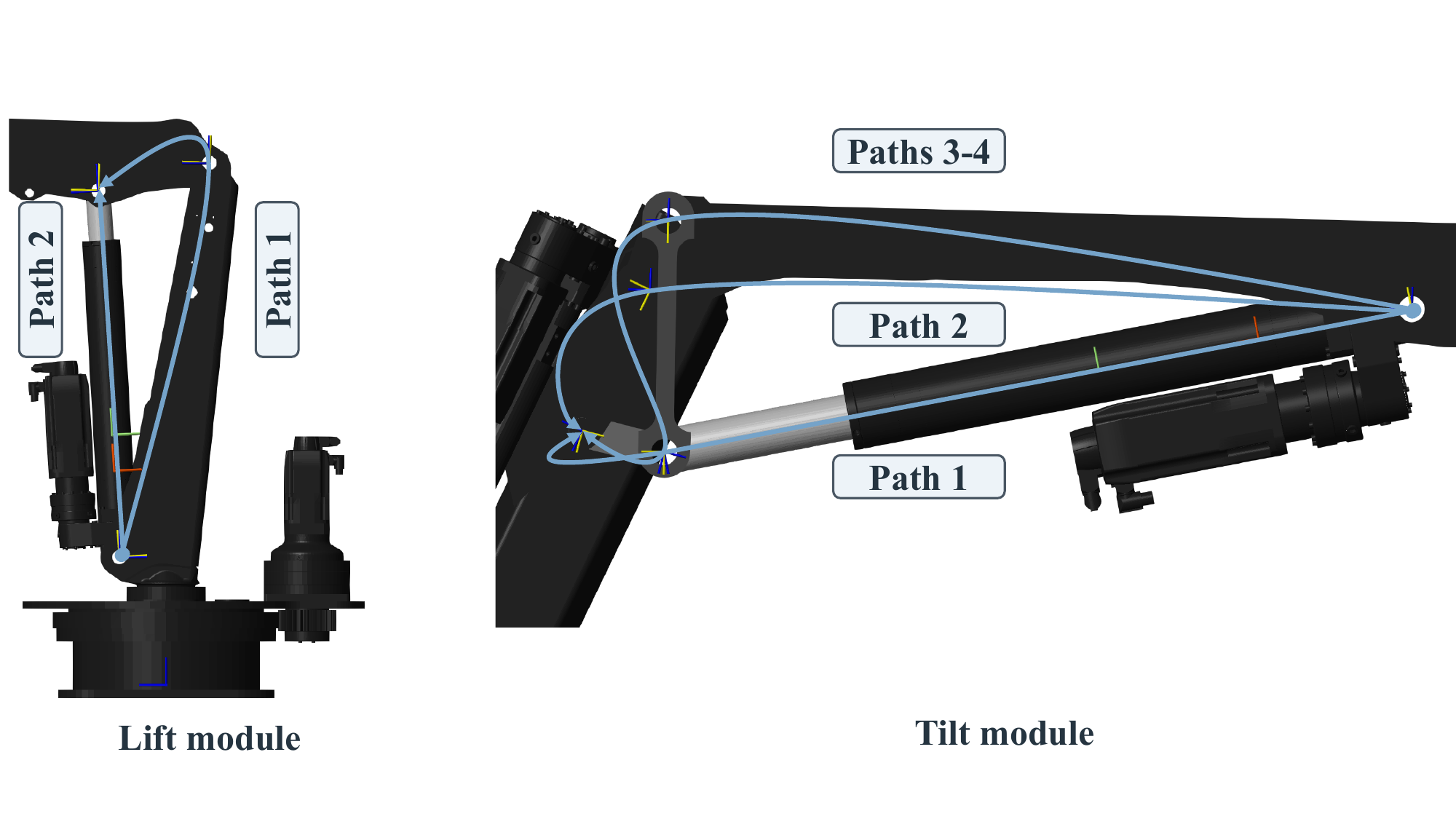}
    \caption{\textit{left}: Manipulator with a four-bar close-up.
    Yellow/gray markers denote active/passive joints.
    The close-up shows the identical
    parallel $a/b$ side links; labels 10 and 12 represent
    $10a/10b$ and $12a/12b$ to avoid clutter.
    \textit{right}: Lift and tilt closed-chain modules.
    Tilt paths 3 and 4 overlap in this side view.}
    \label{fig:robot-modules}
\end{figure*}

\subsection{Diagnostics for Nonunique Reactions}
\label{subsec:reaction-diagnostics}

Nonunique reactions are assessed through the compatibility of
the logged Simscape loads with the rigid-body balance equations
and ideal-joint conditions. These checks are performed after
recovery and do not modify the predicted outputs.

For joint $j$, $\boldsymbol{w}_{c,j}^{\mathrm{SS}}$ denotes
the logged base-on-follower constraint wrench, expressed at
the origin of $F_j$ in its axes.
Projection onto the reaction basis introduced in
Section~\ref{subsec:physical-assembly} gives the reaction
coordinates and excluded wrench component as
\begin{equation}
    \begin{aligned}
        \boldsymbol{\lambda}_j^{\mathrm{SS}}
        &=\boldsymbol{C}_j^T\boldsymbol{w}_{c,j}^{\mathrm{SS}},\\
        \boldsymbol{e}_j
        &=(\boldsymbol{I}_6-\boldsymbol{C}_j\boldsymbol{C}_j^T)
        \boldsymbol{w}_{c,j}^{\mathrm{SS}}.
    \end{aligned}
    \label{eq:reference-joint-projection}
\end{equation}
Joint admissibility is assessed from $\boldsymbol{e}_j$,
with moments and forces normalized by their respective scales,
and the signed quantity
$\boldsymbol{S}_j^T\boldsymbol{w}_{c,j}^{\mathrm{SS}}$.
The latter is the free-axis torque for a revolute joint or
free-axis force for a prismatic joint.

The projected coordinates and logged actuator efforts form
$\boldsymbol{\chi}_{\mathrm{SS}}$ in the load-coordinate
order of~\eqref{eq:assembled-balance}.
The projected balance residual is
$\|\boldsymbol{S}_e^{-1}
(\boldsymbol{A}\boldsymbol{\chi}_{\mathrm{SS}}
-\boldsymbol{b})\|_\infty$.
A second balance restores the excluded components
$\boldsymbol{e}_j$ through the corresponding wrench
transports, with opposite signs on the connected bodies.
This raw balance uses the original logged wrenches and
retains each body's Newton--Euler equations, including
the components removed by projection.
Both balances use the reconstructed motion and therefore
also reflect motion or model mismatch.

The normalized reference vector is
$\boldsymbol{z}_{\mathrm{SS}}
=\boldsymbol{S}_x^{-1}\boldsymbol{\chi}_{\mathrm{SS}}$.
When the balance system is numerically consistent and its
rank is stable, the same orthonormal basis $\boldsymbol{N}$
used in Section~\ref{subsec:output-recovery} gives
\begin{equation}
    \begin{aligned}
        \boldsymbol{\alpha}
        &=\boldsymbol{N}^T
        (\boldsymbol{z}_{\mathrm{SS}}-\boldsymbol{z}_0),\\
        r_{\mathrm{fam}}
        &=\|(\boldsymbol{z}_{\mathrm{SS}}-\boldsymbol{z}_0)
        -\boldsymbol{N}\boldsymbol{\alpha}\|_2 .
    \end{aligned}
    \label{eq:reference-family-distance}
\end{equation}
This is the dimensionless Euclidean distance to the
normalized affine family
$\boldsymbol{z}_0+\operatorname{range}(\boldsymbol{N})$.
Its relative value divides $r_{\mathrm{fam}}$ by
$\max\{\|\boldsymbol{z}_{\mathrm{SS}}\|_2,
\|\boldsymbol{z}_0\|_2,\epsilon_{\mathrm{mach}}\}$,
where machine epsilon $\epsilon_{\mathrm{mach}}$ prevents division by zero.
The norm $\|\boldsymbol{N}\boldsymbol{\alpha}\|_2$ measures
the discrepancy along the numerical nullspace.
The coordinates $\boldsymbol{\alpha}$ depend on the chosen
basis and serve only this diagnostic projection.

A small family distance supports compatibility of the
projected loads with the computed solution family and is
interpreted together with raw balance and joint admissibility.
These diagnostics do not identify physical load sharing
or introduce a compliance model.


\section{Simulation and Results}
\label{sec:results}

\subsection{Manipulator and Simulation Setup}
\label{subsec:simulation-setup}

\begin{figure}[tb]
    \centering
    \includegraphics[page=3,width=0.56\columnwidth,
        trim=0 7mm 0 7,clip]{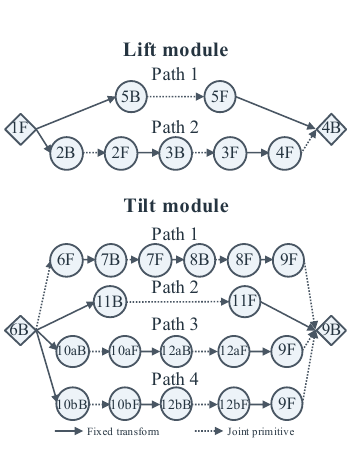}
    \caption{Lift and tilt path graphs. Suffixes $B$ and $F$
    denote base and follower frames; $a/b$ identify the
    parallel links.}
    \label{fig:path-graphs}
\end{figure}

The 7-degree-of-freedom heavy-duty hybrid serial--parallel
manipulator and its path illustrations in
Figs.~\ref{fig:robot-modules} and~\ref{fig:path-graphs} are adapted
from~\cite{dastranj2026}. Geometric and inertial parameters
are protected by a nondisclosure agreement (NDA).
Its tilt four-bar has identical parallel side links with endpoint
joints $(10a,12a)$ and $(10b,12b)$.
Both are modeled separately, giving 14 moving bodies and
18 coordinates: seven active and eleven passive.

The lift and tilt modules have two and four paths, connecting
$1F$ to $4B$ and $6B$ to $9B$, respectively.
Path 1 is the reference, giving one lift and three tilt closure
pairs. Tilt paths 3 and 4 contain the original ($a$) and
additional ($b$) side links, respectively.
The active coordinates are $q_1$, $q_3$, $q_7$, $q_{13}$,
$q_{14}$, $q_{15}$ and $q_{16}$; $q_3$, $q_7$ and $q_{13}$
are prismatic, and all passive coordinates are revolute.

\begin{figure}[tb]
    \centering
    \includegraphics[width=0.90\columnwidth,
        trim=0 0 0 0,clip]{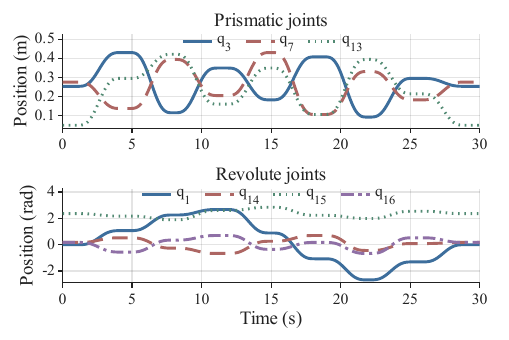}
    \caption{Prescribed active-coordinate trajectories:
    prismatic joints (upper) and revolute joints (lower).}
    \label{fig:trajectory}
\end{figure}

Simscape executes the $30\,\mathrm{s}$ trajectory in
Fig.~\ref{fig:trajectory}, evaluated on a $1\,\mathrm{ms}$
sampling time.
The test includes gravity and zero nongravitational external wrench.
Active positions, velocities and accelerations are the
reconstruction inputs; passive positions and joint loads serve only
for validation.

The assembly has 84 balance equations, 97 load coordinates
and 223 requested scalar outputs. Equation and load-coordinate
scales are $10^4\,\mathrm{N}$ and $10^4\,\mathrm{N.m}$.
The singular-value threshold is
$10^{-12}+10^{-10}\sigma_{\max}$, with $\sigma_{\max}$
the largest singular value of $\boldsymbol{A}_n$.
These settings are independent of the agreement errors.

\subsection{Reconstruction and Load-Recovery Results}
\label{subsec:simulation-results}

Errors are path-assembly values minus Simscape values, with
equally weighted samples for root-mean-square errors (RMSE).
Table~\ref{tab:passive-errors} reports the raw passive-coordinate
errors without angle wrapping. All eleven coordinates are
reconstructed throughout the run; the largest absolute error
is $5.09\times10^{-9}\,\mathrm{rad}$ at $q_8$.
This comparison assesses passive-position reconstruction on the
tested assembly branch.

\begin{figure}[tb]
    \centering
    \includegraphics[width=\columnwidth,
        trim=0 0 0 0,clip]{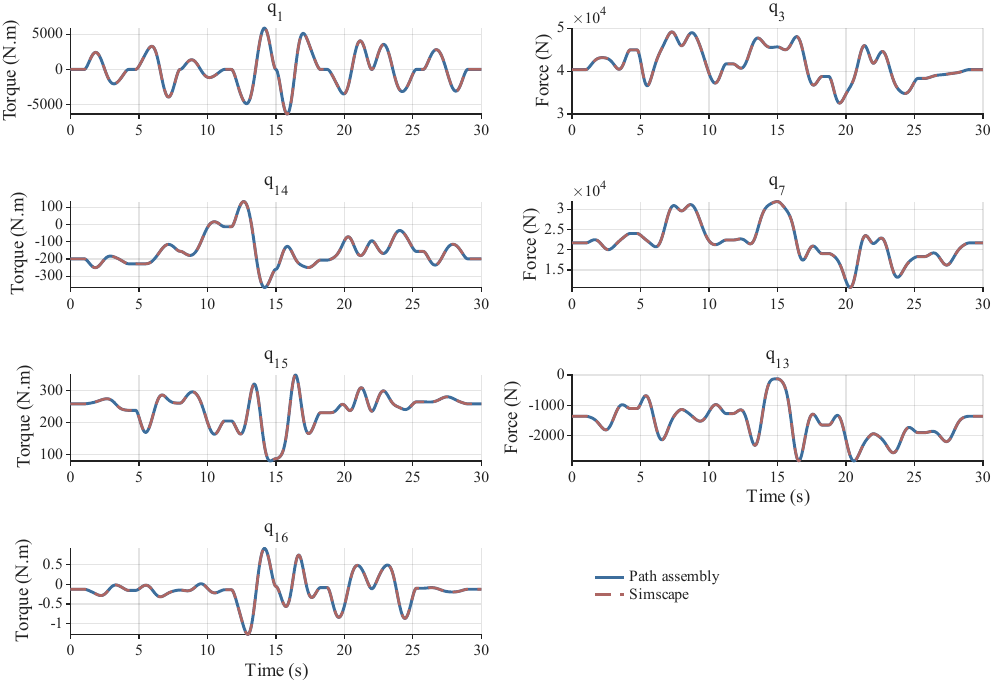}
    \caption{Actuator efforts from path assembly (solid) and
    Simscape (dashed): torques on the left and forces on the right.
    Each joint has an independent vertical scale, shared by the
    two methods; all efforts retain their physical units.}
    \label{fig:actuator-efforts}
\end{figure}

\begin{table}[tb]
\centering\footnotesize
\caption{Passive-coordinate errors (rad), without angle wrapping.
All eleven coordinates have full coverage.}
\label{tab:passive-errors}
\setlength{\tabcolsep}{2pt}
\begin{tabular*}{\linewidth}{@{}l@{\extracolsep{\fill}}rr@{}}
\hline
Joint & RMSE & \shortstack{Max. abs.\\error} \\
\hline
$q_{2}$ & $3.89\times10^{-10}$ & $1.16\times10^{-9}$ \\
$q_{4}$ & $5.97\times10^{-10}$ & $3.01\times10^{-9}$ \\
$q_{5}$ & $8.36\times10^{-10}$ & $4.17\times10^{-9}$ \\
$q_{6}$ & $2.73\times10^{-10}$ & $8.46\times10^{-10}$ \\
$q_{8}$ & $1.21\times10^{-9}$ & $5.09\times10^{-9}$ \\
$q_{9}$ & $1.21\times10^{-9}$ & $4.90\times10^{-9}$ \\
$q_{10a}$ & $3.63\times10^{-10}$ & $2.37\times10^{-9}$ \\
$q_{11}$ & $2.88\times10^{-10}$ & $1.03\times10^{-9}$ \\
$q_{12a}$ & $1.19\times10^{-9}$ & $3.59\times10^{-9}$ \\
$q_{10b}$ & $3.63\times10^{-10}$ & $2.37\times10^{-9}$ \\
$q_{12b}$ & $1.19\times10^{-9}$ & $3.59\times10^{-9}$ \\
\hline
\end{tabular*}
\end{table}

\begin{table}[tb]
\centering\footnotesize
\caption{Actuator-effort errors. All seven outputs are valid at
every evaluated sample.}
\label{tab:actuator-errors}
\setlength{\tabcolsep}{2pt}
\begin{tabular*}{\linewidth}{@{}lc@{\extracolsep{\fill}}rr@{}}
\hline
Joint & Unit & RMSE & \shortstack{Max. abs.\\error} \\
\hline
$q_{1}$ & N.m & $6.32\times10^{-6}$ & $1.41\times10^{-5}$ \\
$q_{3}$ & N & $1.27\times10^{-5}$ & $7.05\times10^{-5}$ \\
$q_{7}$ & N & $1.27\times10^{-5}$ & $8.91\times10^{-5}$ \\
$q_{13}$ & N & $2.24\times10^{-6}$ & $1.16\times10^{-5}$ \\
$q_{14}$ & N.m & $9.68\times10^{-8}$ & $5.06\times10^{-7}$ \\
$q_{15}$ & N.m & $9.96\times10^{-8}$ & $5.04\times10^{-7}$ \\
$q_{16}$ & N.m & $1.64\times10^{-10}$ & $1.55\times10^{-9}$ \\
\hline
\end{tabular*}
\end{table}

\begin{table}[tb]
\centering\footnotesize
\caption{Unique constraint-wrench recovery and errors
relative to Simscape.}
\label{tab:wrench-errors}
\setlength{\tabcolsep}{2pt}
\begin{tabular*}{\linewidth}{@{}lc@{\extracolsep{\fill}}rr@{}}
\hline
Group & \shortstack{Valid/\\requested} & \shortstack{Max. RMSE} & \shortstack{Max. abs.\\error} \\
\hline
Forces (N) & 37/54 & 0.000407 & 0.00113 \\
Moments (N.m) & 28/54 & 0.00641 & 0.00704 \\
\hline
\end{tabular*}
\par\smallskip\raggedright
Valid: uniquely recoverable components satisfying all recovery
checks throughout the run. Errors use these components;
the two maxima may occur in different components.
\end{table}

The balance and rank checks pass at every sample, and all
seven actuator efforts satisfy the recovery conditions.
Figure~\ref{fig:actuator-efforts} compares their histories;
Table~\ref{tab:actuator-errors} quantifies the agreement.
The largest force and torque RMSEs are
$1.27\times10^{-5}\,\mathrm{N}$ and
$6.32\times10^{-6}\,\mathrm{N.m}$, respectively.

Table~\ref{tab:wrench-errors} reports 37 of 54 force components
and 28 of 54 moment components as valid throughout the run.
The requested set comprises all three components at each of
the 18 joints; validity identifies uniquely recoverable outputs
satisfying the checks in Section~\ref{subsec:output-recovery},
independently of their agreement with Simscape.
In a redundant rigid model, individual reactions can vary
within the balance-equation nullspace. Such nonunique components
are assessed through the compatibility diagnostics below.
The counts thus describe recoverability, while the error
statistics measure agreement for uniquely determined components.
The valid counts include structurally zero ideal-joint components.
Each group RMSE is the largest individual component RMSE,
and all error statistics use valid outputs.

\begin{figure}[tb]
    \centering
    \includegraphics[width=0.80\columnwidth,
        trim=0 0 0 0,clip]{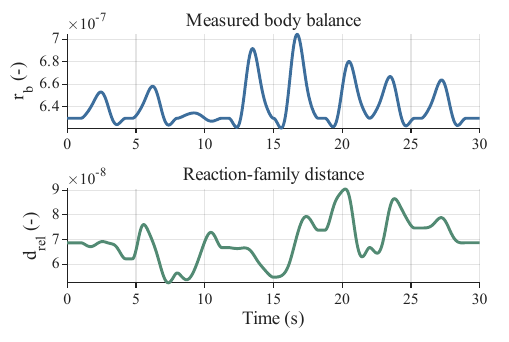}
    \caption{Scaled raw body-balance residual (upper) and
    relative reaction-family distance (lower).
    Both diagnostics are dimensionless.}
    \label{fig:reaction-validation}
\end{figure}

Figure~\ref{fig:reaction-validation} shows the diagnostics
defined in Section~\ref{subsec:reaction-diagnostics}.
Here $r_b$ is the largest absolute scaled residual across the
84 raw body-balance equations, and $d_{\mathrm{rel}}$ is the
relative form of $r_{\mathrm{fam}}$ in that section.
Their temporal RMS values are \mbox{$6.41\times10^{-7}$} and
\mbox{$6.96\times10^{-8}$}, with maxima
\mbox{$7.04\times10^{-7}$} and \mbox{$9.05\times10^{-8}$},
respectively.
The ideal-joint admissibility residual is zero throughout.
These diagnostics support compatibility of the logged
reactions with the numerical load family.

\section{Conclusion}
\label{sec:conclusion}

This work connected analytical path-based motion reconstruction
to physical inverse dynamics and recovery of uniquely determined
loads. The prescribed-motion comparison showed close agreement
for passive positions, all seven actuator efforts and the valid
constraint-wrench components. The reaction diagnostics supported
compatibility with the admissible numerical load family without
selecting a unique redundant reaction distribution. These findings
apply to the tested regular branch and rigid-body, ideal-joint
model; compliant load sharing remains outside the formulation.


\section*{Acknowledgements}
We acknowledge the financial support of the Finnish Ministry of Education and Culture through the Intelligent Work Machines Doctoral Education Pilot Program (IWM VN/3137/2024-OKM-4).


\bibliographystyle{IEEEtran}
\bibliography{ref}

\end{document}